\documentclass{eptcs}

\usepackage[T1]{fontenc}
\usepackage[sf,scaled=0.85]{noto}
\usepackage[scaled=0.85]{GoMono}
\usepackage{amssymb}

\usepackage{theorem}
\usepackage{mytheorems}

\usepackage[fleqn]{mathtools}
\usepackage{parskip}
\usepackage{hyperref}
\usepackage{xspace}
\usepackage{subfigure}
\usepackage{graphicx}

\graphicspath{{figures}}

\usepackage[frozencache]{minted}
\newmintinline[fm]{prolog}{fontsize=\small,showspaces=true,space=\;}

\setminted[prolog]{
  fontsize=\small,
  baselinestretch=1.05,
  mathescape=true,
  showspaces=true,
  space=\;
}

\usepackage{pseudo}
\pseudoset{
ct-left={{~\small//~}},
ct-right=\relax,
indent-length=1em,
label={\small\arabic*},
kw,
}

\RequirePackage{tabularx}%
\newcolumntype{L}{>{$}l<{$}}%
\newcolumntype{C}{>{$}c<{$}}%
\newcolumntype{R}{>{$}r<{$}}%
\newcolumntype{M}{@{}p{\mathindent}@{}}%
\newcolumntype{t}{>{\mbox\bgroup}l<{\egroup}}%
\newcolumntype{e}{r@{\;}l}%
\newcolumntype{E}{>{$}r<{$}@{$\;$}>{$}l<{$}}%

\def\funfont{\upshape\ttfamily}
\def\fun#1{\text{\funfont #1}}

\newcommand{\CM}{\mathbin{\fun{\small :--}}}
\newcommand{\QM}{\mathop{\fun{ ?--}}\;}
\newcommand{\CC}{\mathbin{\fun{\small::}}}
\newcommand{\AT}{\mathbin{\fun{@}}}
\newcommand{\TT}{\mathit{tt}}
\newcommand{\PR}{\mathit{pr}}
\newcommand{\PLUS}{\mathbin{\fun{+}}}
\newcommand{\PLUSPLUS}{\mathbin{\fun{++}}}
\newcommand{\MINUSMINUS}{\mathbin{\fun{--}}}
\newcommand{\SIM}{\mathbin{\fun{$\sim$}}}
\newcommand{\EQ}{\mathbin{\fun{=}}}
\newcommand{\LB}{\fun{[}}
\newcommand{\RB}{\fun{]}}
\newsavebox{\notBox}
\begin{lrbox}{\notBox}\verb|\+|\end{lrbox}
\newcommand{\NOT}{\mathop{\usebox{\notBox}}}

\newcommand{\Prob}{\mathrm{P}}

\newcommand{\q}{\mathrm{q}}

\DeclareMathOperator{\strat}{\mathit{strat}}
\DeclareMathOperator{\reg}{\mathit{reg}}
\DeclareMathOperator{\gnd}{\mathit{gnd}}
\DeclareMathOperator{\var}{\mathit{var}}

\DeclareMathOperator{\fvar}{\mathit{fvar}}

\DeclareMathOperator{\dom}{\mathit{dom}}

\usepackage{xcolor}

\newcommand{\Fusemate}{\textsf{Fusemate}\xspace}

\definecolor{DodgerUniformBlue}{rgb}{0.0,0.353,0.612}
\newcommand{\define}[1]{\emph{\textcolor{DodgerUniformBlue}{#1}}}

\definecolor{paleyellow}{HTML}{FFEC7F}

\long\def\comment[#1]#2{\par\colorbox{paleyellow}{\llap{\textbf{#1:\quad}}%
    \parbox[t]{\textwidth}{\setlength{\parskip}{1ex plus 0.2ex minus 0.2ex}#2}}}

\def\void{}
\newcommand{\mcomment}[2][\void]{\marginpar{\raggedright\footnotesize\ifx\void#1\else\textbf{\tiny
        #1:\\}\fi#2}}

\usepackage{url}
\usepackage{natbib}
\setcitestyle{numbers,square,comma} 

\usepackage{old-arrows}
\newcommand{\ce}{\coloneq}

\title{Bottom-Up Grounding in the Probabilistic Logic Programming System Fusemate}
\author{Peter Baumgartner
\institute{Data61/CSIRO\\ Canberra, Australia}
\email{Peter.Baumgartner@data61.csiro.au}
\and
Elena Tartaglia
\institute{Data61/CSIRO\\ Melbourne, Australia}
\email{Elena.Tartaglia@data61.csiro.au}
}

\begin{document}
\maketitle

\begin{abstract}
  This paper introduces the \Fusemate probabilistic logic programming  
  system. \Fusemate's inference engine comprises a grounding
  component and a variable elimination method for probabilistic inference.
  \Fusemate differs from most other systems by grounding the program
  in a bottom-up way instead of the common top-down way.  
  While bottom-up grounding is attractive for a number of reasons, e.g., for dynamically
  creating distributions of varying support sizes, it makes it harder to control the amount of ground clauses
  generated. We address this problem by interleaving 
  grounding with a query-guided relevance test which 
  prunes rules whose bodies are inconsistent with the query. 
  We present our method in detail and demonstrate it with examples that involve ``time'', such as (hidden) Markov models.
  Our experiments demonstrate competitive or better
  performance compared to a state-of-the art probabilistic logic programming
  system, in particular for high branching problems.\footnote{This is an extended version
    of the ICLP 2023 paper
    at~\url{https://cgi.cse.unsw.edu.au/~eptcs/paper.cgi?ICLP2023:4654}.
    It also includes an improvement to the grounding algorithm in Section~\ref{sec:grounding}.
  }
  \end{abstract}

 

\section{Introduction}
\label{sec:introduction}
Probabilistic Logic Programming (PLP) combines logic programming with probability theory.
Most PLP systems implement the \emph{distribution semantics}
(DS)~\cite{sato_statistical_1995}. The DS was introduced for ground definite programs
whose facts are annotated with probabilities.
More expressive languages 
require additional concepts to equip their programs with a DS. 
ProbLog~\cite{fierens_inference_2015}, for instance, features
probability annotations of rule heads, not just facts, 
disjunctive heads, for expressing distributions, and an expressive form of default
negation. These are dealt with by a combination of program transformation, and defining the DS via two-valued well-founded models
of the grounded transformed program (if it exists)~\cite{de_raedt_probabilistic_2015}.
After grounding, 
a probabilistic inference engine solves a given inference problem, e.g., the
probability of a query being satisfied in the models of the program.
Many PLP systems implement their grounding component (or monolithic reasoner) in a top-down way.

In this paper, we investigate the potential of bottom-up grounding as an
alternative to the more common top-down grounding. 
Our main motivation is efficient reasoning in applications with a high branching
rate, for approximating continuous distributions, in a timed setting (Markov
models). Our main results are a novel method for bottom-up
grounding, its implementation in our PLP system \Fusemate, and the favourable
experimental results we obtained with it.
A main challenge with bottom-up grounding is controlling the amount of ground rules
generated. Our method addresses this problem by 
pruning redundant rules based on a certain inconsistency test. 
The test is applied dynamically, at each step of the grounding process along a certain
program stratification ordering. Our method supports an expressive input language with
default negation, where default negation is eliminated on-the-fly at each step.

In the main part of the paper, we describe the method in detail and prove its correctness.
The rest of this introduction has more motivation, overviews the main ideas, and discusses
related work.






\paragraph{Bottom-up grounding.}
We propose a method for grounding the program in a bottom-up way as an alternative to the
top-down approach described above.  Our grounding is defined for a certain form of program
stratification called \emph{SBTP}, defined in Section~\ref{sec:stratified-logic-programs}
below.  The result ground program is obtained via fixpoint computation along this
stratification order. In each step of the fixpoint computation the program rules of the proper stratification
level are grounded thereby also eliminating occurrences of default negation.
This is advantageous for on-the-fly removal of redundant ground rules, which
can be informed with the result program lower than the stratum of the current step.
See below for details.

The default negation operator is expressive and
allows for negation over (implicitly) existentially quantified conjunctions of atoms.  
As an example consider this program
(\Fusemate uses ProbLog syntax):
\begin{minted}{prolog}
0.5 :: q(0).    0.5 :: q(1).    0.5 :: q(2).    0.5 :: p(T) :- q(T), \+ (q(S), S < T). 
\end{minted}
Its grounding has the same three facts and the
rule above replaced by the following three ground rules:
\begin{minted}{prolog}
0.5::p(0) :- q(0).    0.5::p(1) :- q(1), -q(0).    0.5::p(2) :- q(2), -q(0), -q(1).
\end{minted}
These rules explain all ways for \fm{p}-literals to become true in the least Herbrand model extending
a given choice of the probabilistic facts, where  $\text{\fm{-}}a$ is true if the fact 
$a$ is false. They are computed bottom-up with a stratification that puts
$\text{\fm{q}}$-atoms before $\text{\fm{p}}$-atoms.
With this approach we obtain an extension of the DS that is compatible with, e.g.,
the transformation-based approach of ProbLog mentioned above.

\paragraph{Motivation.} We are interested in exploring the potential of bottom-up grounding for practical
applications.
An example are \emph{dynamic distributions}, i.e.,
distributions whose domain is dynamically altered by the program.
Dynamic distributions are not supported by most PLP systems (exceptions are, e.g.,
CP-logic~\cite{vennekens_cp-logic_2009}, and the approach by ~\citet{gutmann_magic_2011}). 
\Fusemate 
has a construct $t \SIM l$ where $t$ is a term for a random 
variable and $l = \LB v_1,\ldots, v_n \RB$ is a Prolog list of values to draw from, uniformly
by default.
The list can be constructed by the program and draw cases can be queried as equations $t \EQ v_i$.
The following program models
iterated drawing balls from an urn without replacement
(an annotation $a\AT T$ stands for a dedicated
``time'' argument $T$ considered part of atom $a$'s arguments; list append $\PLUSPLUS$ and difference
$\MINUSMINUS$ are built-in infix operators):
\begin{minted}{prolog}
urn([r(1), r(2), g(1)]) @ 0.               %% Initially two red and one green distinguishable balls
draw ~ Balls @ T :- urn(Balls) @ T, Balls \= [].      %% Draw  a ball uniformly if urn is not empty
urn(Balls -- [B]) @ T+1 :- urn(Balls) @ T, draw = B @ T.     %% Drawing a ball removes it from urn
some(red) @ T :- draw=r(_) @ T.                                                    %% Abstract from ball id, color only
some(green) @ T :- draw=g(_) @ T.
\end{minted}
To give an idea of the overall system, \Fusemate can solve queries like the following:
\begin{minted}{prolog}
?- some(green) @ 0.  % 0.333333
?- some(green) @ 1 | some(red) @ 0. % 0.5 conditional query
?- some(C1) @ 1, some(C2) @ 2 | some(red) @ 0. % Non-ground conditional query, two solutions:
% 0.5 :: [C1 = red, C2 = green], 0.5 :: [C1 = green, C2 = red]                   
\end{minted}

As in the example above, we are particularly interested in applications that involve 
``time'', such as hidden Markov models and their enrichments with rules for
domain constraints.
Such applications are within the scope of many PLP systems,
see~\cite[e.g.]{hutchison_logic_2004,sato_prism_1997}. In
Section~\ref{sec:application-examples} below we report on experimental results on Markov models
which demonstrate the viability of our bottom-up approach.\footnote{Bottom-up grounding
  also makes it easy to  catch positive loops.
\Fusemate deals with cyclic programs by means of certain query rewriting
techniques. The details are beyond the scope of this paper.}




\paragraph{Query-guided grounding and inconsistency pruning.}
The advantages of  bottom-up grounding are offset by some drawbacks.
A major drawback is its inherent lack of
goal-orientedness. Compared to the satisfiability problem for
(stratified) non-probabilistic programs, the problem is amplified in a probabilistic
setting as exponential branching into model candidates is unavoidable. 
To address this problem we developed a technique for \emph{query-guided} bottom-up
grounding. It has two components which reinforce each other:
\emph{goal regression} and \emph{inconsistency pruning}. They rest on a fixed
interpretation of equations as \emph{right-unique relations}, which is also taken
advantage of in the inference component.

To explain, let a query be a set of ground literals.
Goal regression makes the query more constrained by adding literals to it.
To ensure models are preserved, only literals that are included in \emph{every} model of the query are added.
 This is done iteratively along the stratification order and using the
grounded rules so far at each step. Inconsistency pruning prevents, at each step in the iteration,
applying rules for grounding that cannot contribute to a model of the query.
As an example consider this Hidden Markov Model:
\begin{minted}[linenos]{prolog}
state ~ [[rainy, 0.6], [sunny, 0.4]] @ 0.   %% T = 0, non-uniform distribution given
obs ~ [3..30] @ 0 :- state=rainy @ 0. %% Observation is accumulated rainfall over time in mm
obs ~ [0..5] @ 0 :- state=sunny @ 0.
state ~ [[rainy, 0.7], [sunny, 0.3]] @ T+1 :- state=rainy @ T.   %% T -> T+1
state ~ [[rainy, 0.4], [sunny, 0.6]] @ T+1 :- state=sunny @ T.
obs ~ [R+3..R+30] @ T :- state=rainy @ T, T > 0, obs=R @ T-1. %% Increase observation at T-1
obs ~ [R..R+5] @ T :- state=sunny @ T, T > 0, obs=R @ T-1.
\end{minted}
Take the query \fm{?- obs=0 @ 0, obs=4 @ 1, obs=20 @ 2, obs=24 @ 3.} The task is to compute its success probability.
Note that an increase of \fm{obs}erved rain by $0$ at time \fm{0} can be proven
\emph{only} through the subgoal \fm{state=sunny@0} as per the
rule on line 3. Goal regression detect this and extends the query with \fm{state=sunny@0}.
Then, inconsistency pruning applies and rejects the rule on line 2 because its body literal
\fm{state=rainy@0} is \emph{inconsistent} with the now extended query. This inconsistency
follows from our fixed semantics of equations for representing functions, i.e.,
right-unique relations. With that, \fm{state=sunny@0} and \fm{state=rainy@0} are
inconsistent  as \fm{sunny}  $\neq$ \fm{rainy}, by unique name assumption.
The ground instance \fm{state ~ [[rainy, 0.7], [sunny, 0.3]] @ 1 :- state=rainy @ 0} is
rejected for the same reason. Also, any instance of the rule on lines 6 or line 7 with
$\text{\fm{T}} = 1$ and $\text{\fm{R}} \neq 0$ will have its body literal \fm{obs=R @ T-1}
inconsistent with the query literal \fm{obs=0 @ 0}.
The justification for inconsistency pruning is that no model of the query can satisfy the body
of a rule that is inconsistent with the query. Rejected 
rules can only ``fire'' in non-models of the query. As these interpretations receive
probability 0 anyway, such rules can just as well be ignored.
Goal regression and inconsistency happen during each step
of the bottom-up grounding process.
The general technique is detailed in
Section~\ref{sec:grounding}. 





\paragraph{Related work.}
Too many PLP systems have been developed to discuss them
all in detail. \citet{de_raedt_probabilistic_2015}, and
\citet{riguzzi_survey_2017} provide overviews.
\citet{fierens_inference_2015} explains ProbLog  in depth and
compares it with related systems. 
Systems with built-in top-down grounding are 
based on variants of SLD-resolution (e.g., 
ProbLog, PITA~\cite{riguzzi_pita_2011}) and construct (ground-level) stratified programs by design.
The most closely related work comes from systems with built-in bottom-up grounding and
that optimize the grounding in some way. This excludes systems that assume an external
grounding component but otherwise have a related execution model, e.g.,
CP-Logic~\cite{vennekens_cp-logic_2009}, which also supports dynamic   
distributions.
\citet{vlasselaer_t_2016} describe a fixpoint approach for incrementally grounding
stratified programs with possibly positive cycles. Their grounding works incrementally
and supports any-time reasoning for varying queries. 
Grounding can be controlled  heuristically, which guarantees lower success probabilities
bounds, or,  in certain cases, exact probabilities, e.g., for filtering queries in Hidden Markov Models.
Several authors have considered (variants of)  magic set transformations for controlling
the grounding process. The idea is to avoid bottom-up inferences with program rules that
are impossibly relevant for proving 
the query at hand. Relevance is determined in a regression process from the query towards
the facts using the program rules. \citet{gutmann_magic_2011} introduced magic sets for
probabilistic logic programs. Their approach supports dynamic distributions but not default negation.
\citet{tsamoura_beyond_2020} extend the work by \citet{vlasselaer_t_2016} by adding
an incremental magic set transformation during grounding.
While the goal of magic sets  and our inconsistency pruning are similar, their realizations are different. 
 Our inconsistency pruning also regresses the query, however, by
adding literals in the \emph{intersection} of all rule applications instead of taking unions. Furthermore, pruning
is based on \emph{inconsistency} of grounded bodies with the query, which is not the case in
magic sets. The HMM example above, for instance, is not covered by magic sets.
As future work, it seems promising to combine these orthogonal techniques.
%
%
The first author introduced SBTP in an earlier \emph{non-probabilistic}
Fusemate system~\cite{baumgartner_possible_2020,baumgartner_fusemate_2021} and combined it
with description logic 
reasoning and the event calculus~\cite{baumgartner_combining_2021}.
In this paper we adapt SBTP to a probabilistic setting and integrate query guidance into
it. The unguided grounding in these earlier \Fusemate systems is unsuitable for all but
the smallest problems.
\citet{skarlatidis_probabilistic_2015} use ProbLog for event calculus reasoning, which is
an interesting application for us as well.

\section{Stratified Logic Programs}
\label{sec:stratified-logic-programs}
Assume as given a first-order logic signature comprised of function and predicate symbols of
fixed arities. Assume a countable set of variables.
A \define{term} is either a variable or an expression of the form $f(t_1,\ldots,t_n)$ where $f$ is a function symbol and
$t_1,\ldots,t_n$ are terms. Atoms are of the form $p(t_1,\ldots,t_n)$ where $p$ is a
predicate symbol. 
A \define{ground} term or atom does not contain any variable.
We also use vector notation $\vec t$ for lists
$t_1,\ldots,t_n$ of objects, where $n \ge 0$. 
We distinguished between ordinary (free) symbols and ``built-in'' symbols. The latter
include the integer constants $\mathbb{Z}$ and infix operators  like $<$ and $+$.
A term or atom with a built-in symbol is called an
\define{interpreted} term or atom, otherwise it is \define{ordinary}. We require that
all terms are well-sorted, so that interpreted ground terms can
effectively be evaluated to a unique irreducible term. Interpreted ground atoms must
be valuable to a Boolean $\top$ or $\bot$. 
lists as in Prolog.
%
Every ordinary atom must have a dedicated $\mathbb{Z}$-sorted argument,
say, the first one, its \define{time term} $\TT$. When we write $p(\vec t) \AT \TT$ we mean 
$p(\TT, \vec t)$ instead. If $a$ is an atom we write
$a \AT \TT$ to indicate it has time term $\TT$, i.e., $a$ is of the form  $p(\TT, \vec t)$.
We assume an ordinary ternary infix predicate
symbol $\EQ$ and write $\EQ$-atoms as $s \EQ t \AT \TT$. An $\EQ$-atom is
\define{admissible} if its left hand side term $s$ is an ordinary functional term $f(\vec
t)$. We only work with admissible $\EQ$-atoms.
An admissible $\EQ$-atom $f(\vec t) \EQ t \AT \TT$ could be taken as an
ordinary atom $f(\vec t,t) \AT \TT$ with $f$ as a predicate symbol.
%
Let $e$ be a term or atom. By $\var(e)$ we denote the set of variables occurring in $e$.
Let $\sigma$ be a substitution. We write
$e\sigma$ for the term or atom resulting from applying $\sigma$ to $e$. The domain of $\sigma$ is 
$\dom(\sigma)$.  A substitution $\gamma$ is a \define{grounding substitution for a set of
  variables $X$} iff $\dom(\gamma) = X$ and $x\gamma$ is ground for every
$x\in\dom(\sigma)$. 
In the following,  the letters $x,y,z$ stand for variables, $p,q,r$ for predicate symbols, and $s, t$ for terms, possibly
indexed. 
A \define{(Fusemate) rule} is an implication of the form 
    \begin{equation}
      \label{eq:rule}
      H \CM b_1,\ldots,b_k, \NOT \vec c_1, \ldots ,\NOT \vec c_n \enspace.
    \end{equation}
where  $b_1,\ldots,b_k$ is a sequence of atoms and each $\vec c_j$ is a sequence of atoms, for
some $k, n \ge 0$. The part to the right of $\CM$ is called the \define{body} $B$.
Every $b_i$ is called a \define{positive (body) literal} and every $\NOT \vec c_j$
is called a \define{negative body element}. A singleton $\NOT \vec a$ is also called a
\define{negative (body) literal} and also written as \define{$\neg a$}. 
The \define{head $H$} is one of the following: (a) an \define{ordinary} head $\PR \CC a
\AT \TT$, (b) a \define{distribution} head $f(\vec t) \SIM t \AT \TT$, or (c)
a \define{sum} head $\PR_1 \CC a_1 \AT \TT \PLUS \cdots \PLUS \PR_m \CC a_m \AT \TT$, for some $m \ge 2$.
Here, $a_{(i)}$ are ordinary atoms, $t$ is an ordinary term, $\PR_{(j)}$ are variables
or built-in real-sorted terms, and
$\TT$ is the \define{time term (of the rule)}.
We write $a \AT \TT$ as short form for the ordinary head $1.0 \CC a \AT \TT$.

A body $B$ is \define{variable-free} iff $X_B \ce \fvar(b_1,\ldots,b_k) = \emptyset$. 
A rule is \emph{range-restricted} if $\fvar(H)  \subseteq X_B$. 
A range-restricted rule is \define{variable-free} iff $X_B = \emptyset$. It is
\define{ground} iff additionally $\fvar({\vec c_i}) = \emptyset$ for each $\vec c_i$.
(In variable-free rules, the extra variables in the $\vec c_i$ are implicitly existentially quantified.)
A \define{fact} is a rule with $k,n = 0$, still matching ``$H \CM B$'' where $B$ is empty,
but also written as $H$. 
Notice that range-restricted facts are always ground. In the following we work with range-restricted rules only.
A ground body $B$ is \define{normal} if all its negative body elements are literals,
i.e., $B$ has the form $b_1,\ldots,b_k, \neg c_1,\ldots,\neg c_n$ where all $c_i$ are ordinary atoms.
A set of ground rules is \define{normal} iff all non-fact rules have a normal body and an
ordinary head with probability $1$. Facts with a head probability $< 1.0$ are also called
\define{probabilistic facts}.


The standard notion of stratification~\cite{apt_towards_1988} is equivalent to saying that the call graph of
a normal logic program has no cycles going through negative body literals.
Every strongly connected component of the call graph is called a stratum and contains the
predicates that are defined mutually recursive with each other.  
In stratified programs, negative body literals can only contain atoms with predicates
defined in lower strata.
The semantics of stratified programs is defined by bottom-up model computation on the
program's ground instances from lower to higher strata until fixpoint. In that, rule
heads are inserted into the model only if their body is true.
Importantly, the truth value of a negative body literal 
depends alone from the models at the lower strata, which have all been
computed in full earlier.

\Fusemate employs a weaker notion of stratification that we call \define{stratification by time and by
  predicates (SBTP)}~\cite{baumgartner_fusemate_2021}. SBTP generalizes standard stratification
to a lexicographic combination of the standard ordering on time (the non-negative integers) and standard
stratification.
For the purpose of defining SBTP, every $\EQ$-atom $f(\vec t) \EQ t \AT \TT$ or
distribution head $f(\vec t) \SIM t \AT \TT$ is taken as an ordinary atom with
\emph{predicate} symbol $f$. 
We say that a set of rules is \define{time constrained} if
for every non-fact rule \eqref{eq:rule} there is a variable $n$
(for ``now'') that is the time term of a positive body literal $b_i$ and such that
\begin{itemize}
\item[(i)] the time term of every other ordinary positive body atom $b_j$ is constrained to $\ge n$.
\item[(ii)] the time term of $H$ is either (a) the variable $n$ or else (b) constrained to $>  n$.
\item[(iii)] the time term of every ordinary atom $b$ in every $c_j$ is 
  constrained to either  (a) $\ge  n$ or else (b) $>  n$.
\end{itemize}
Every such atom $b_i$ is called a \define{pivot (of the rule)}.\footnote{Pivots are not
  needed in variable-free rules as the head can be used in lieu. For example, a
  rule \fm{all_innocent @ 0 :- \+ guilty(Person) @ 0} can be
  accommodated by taking \fm{all_innocent @ 0} as the pivot.  
}
To time constrained  sets of rules we associate a call graph that ignores all rules satisfying 
(ii)-(b) and in all remaining rules ignores all ordinary atoms $b$ in all $c_j$ that
satisfy (iii)-(b). Intuitively, (ii)-(b) rules define an atom in the future and, hence, are
irrelevant for time-stratified model computation at  ``now''; the (iii)-(b) atoms are all
computed strictly earlier than ``now'' and this way are already time-stratified.
(All other cases need a second-tier call graph based stratification.)
A predicate symbol of an ignored
atom or an atom in an ignored rule is added as an isolated node to the call graph if it
does not already occur in it. 
We say that a set of rules \define{is SBTP} if it is time-constrained and its 
call graph has no cycles through negative body elements.  
A \define{(Fusemate) program $P$} is a finite set of range-restricted rules that is SBTP.
The \define{stratification} induced by the call graph is a set
$\{s_1, \ldots, s_m\}$ of sets of predicates equipped with a strict partial order $\prec$
(``lower-than'') induced by the edges in the call graph. Every $s_i$ is a strongly
connected component in the call graph. We identify $\prec$ with an arbitrary extension
to a linear order $s_1 \prec \cdots \prec s_m$.
A \define{timed stratum} is
a pair $(n, s_k)$ where $n \ge 0$ and $1 \le k \le m$. We extend $\prec$ 
lexicographically to timed strata, also denoted by $\prec$.
If  $a$ is a ground ordinary atom $p(\vec t)  \AT n$ define $\define{$\strat(a)$} = (n, s_k)$ where $s_k \ni p$.

For example, a program rule 
\fm{p @ N :- q @ N, \+ (r @ T, T <= N)} has to be predicate stratified with
$\fun{r} \prec \fun{q}$ to be SBTP. 
The rule \fm{p @ N :- q @ N, \+ (r @ T, T < N)} can have $\fun{r}$ and $\fun{q}$ unrelated
because \fm{T < N} already provides time constrained stratification.
We found the two-level stratification useful in practice. It helps, for instance, to embed the fluent
calculus. See~\cite{baumgartner_combining_2021,baumgartner_possible_2020} for more motivating usages.



\section{Bottom-Up Grounding and Distribution Semantics}
\label{sec:grounding}
In this section we define our bottom-up grounding, define a Distribution semantics for
\Fusemate programs, and prove the correctness of our pruning technique.

Let $D$ be a set of ordinary atoms, the \define{domain}.
A \define{(Herbrand) interpretation $I$ over $D$} is any subset $I \subseteq D$. The atoms in $I$
are ``true'' and those in $D \setminus I$ are ``false''. 
We define a satisfaction relation $\models$ as follows: 
for ground ordinary atoms $a$, $I \models a$ iff $a \in I$
for ground built-in atoms $a$, $I \models a$ iff $a$ evaluates to $\top$;
for sequences $\vec a$ of ground atoms, $I \models \vec a$ iff $I\models a$ for all $a \in \vec a$.
A \define{matcher} for a sequence $\vec a$  of atoms to $I$ is a grounding
substitution $\gamma$ for $\fvar(\vec a))$ such that $I \models \vec a\gamma$. 
For a negative body element,  $I \models \NOT \vec c$ iff there is no matcher for $\vec c$
to $I$.
Notice that for ground ordinary atoms $a$,  $I \models \neg a$ iff $a \notin I$ iff $I \not\models a$.


The following procedure eliminates negative body elements from a variable-free body $B$ by 
introducing ground negative body literals. Parameterized in a domain $D$, it returns a
(disjunctive) set of bodies that is equisatisfied with $B$ for any interpretation $I$ with
domain $D$, see Lemma~\ref{lemma:grounding} below.
\begin{procedure}[Body grounding $\gnd_D(B)$]
\label{proc:grounding}
Let $B$ be a variable-free body and $D$ a set of
ground ordinary atoms. Initiate a result set $R \ce \{ B\}$. As long as $R$ contains a body
$B', \NOT \vec c$ with $\NOT \vec c$ not a ground negative body literal,
 do the following: 
let $E = \{ \vec c\gamma_1, \ldots,\vec c\gamma_k \}$ where $\gamma_1,\ldots,\gamma_k$ are all matchers of
$\vec c$ to $D$ ($k=0$ is possible). 
Let $H$ be the set of all hitting sets of $E$.\footnote{Given a set $E = \{E_1,\ldots,E_n\}$,
  $n\ge 0$, where each
  $E_i$ is a sequence of  elements,  a hitting set $h$ of $E$ is a subset-minimal set that contains
  an element from each $E_i$. If $E= \emptyset$ then uniquely $h = \emptyset$.} 
$H$ determines explanations for making the  $\vec c\gamma_i$ false in $D$, as follows:
let $R' = \{ B',\neg d_1,\ldots,\neg d_k\mid \{d_1,\ldots,d_k\} \in H\}$ and set
$R \ce (R \setminus  \{B', \NOT \vec c\}) \cup R'$. 
After completion, $R$ is a set  of ground bodies with negative body literals only.
Finally apply obvious Boolean simplification rules to $R$, enabled by the constants $\top$ and
$\bot$ present after evaluating built-ins. 
The resulting set is denoted by \define{$\gnd_D(B)$}.
\end{procedure}
\begin{example}[Grounding]
\label{ex:grounding}
Consider the following program (time is implicitly 0).
\begin{minted}{prolog}
0.5 :: p(a).        q(a).        0.5 :: p(c).         
0.5 :: p(b).        q(b).        s :- \+ (p(X), q(X)).
\end{minted}
Let 
$D = \{\text{\fm{p(a)}}, \text{\fm{p(b)}}, \text{\fm{q(a)}}, \text{\fm{q(b)}}, \text{\fm{p(c)}}\}$.
The grounding of the body $\gnd_D(\text{\fm{\+ (p(X), q(X))}})$ has four bodies
\fm{(¬q(a), ¬q(b))}, 
\fm{(¬q(a), ¬p(b))}, 
\fm{(¬p(a), ¬q(b))}, and
\fm{(¬p(a), ¬p(b))}.
They represent all four ways to make the body true in
interpretations $I \subseteq D$. 
Notice that $\{\text{\fm{p(c)}}, \text{\fm{q(c)}}\}$ is not in $E$ and hence
is ignored for grounding. 
\end{example}
Lemma~\ref{lemma:grounding} states the core property for correctness of grounding
(proofs are in the Appendix).

\begin{lemma}[Grounding preserves semantics]
\label{lemma:grounding}
Let $D$ a domain, $I \subseteq D$ and $\vec c$ a sequence of atoms.
Let $E = \{ \vec c\gamma_1, \ldots,\vec c\gamma_k \}$ where $\gamma_1,\ldots,\gamma_k$ are all matchers of
$\vec c$ to $D$ ($k=0$ is possible) and $H$ the set of all hitting sets of $E$.
Then $I \models \NOT \vec c$ iff there is a $\{d_1,...,d_k\} \in H$ such that $I \models \neg d_1,...,\neg d_k$.
\end{lemma}



The following function \pr{Normal}(r) expands a ground rule with a sum head into normal form.
It splits disjunctions into disjoint cases using default negation, and extracts
head probabilities by introducing probabilistic facts. This technique is well-known~\cite{poole_probabilistic_1993}.
\begin{pseudo}*
\hd{Normal}(\PR_1 \CC a_1 \AT\TT\PLUS \cdots \PLUS \PR_m \CC a_m \AT\TT \CM B) \ct{$m \ge 1$,
  possibly empty $B$} \\
$M := 1.0$ \ct{probability mass available for remaining cases} \\
$G \ce \{\mathrm{head}_B \AT \TT \CM B \}$ \ct{result set of rules, introduce name $\mathrm{head}_B\AT\TT$ 
  for head}\\
for $i \gets 1 \.. m$ do \\+
$G \ce G \cup \{ a_i \AT \TT \CM \mathrm{head}_B \AT \TT, \neg\mathrm{case}_{B,1} \AT \TT,\..,\neg\mathrm{case}_{B,i-1}\AT \TT,\mathrm{case}_{B,i}\AT \TT \}$\\
$G \ce G \cup \{(M \cdot \PR_i) \CC \mathrm{case}_{B,i} \AT \TT\}$\\
$M \ce M \cdot (1.0 - \PR_i)$\\-
return $G$
\end{pseudo}
\pr{Normal} is applied to ground rules with distribution heads $f(\vec t) \SIM \LB t_1,\ldots,,t_m \RB \AT \TT$, where
$\LB t_1,\ldots,,t_m \RB$ is a non-empty term list, by prior conversion to the sum head 
$1/m \CC f(\vec t) \EQ t_1 \AT \TT  \PLUS \cdots \PLUS 1/m \CC f(\vec t) \EQ t_m \AT \TT$.



To define bottom-up grounding,
let $B$ be a body as in \eqref{eq:rule},
$D$ a set of ordinary atoms, and $S$ a timed stratum.
A matcher $\gamma$ for $b_1,\ldots,b_k$ to $D$ is called an \define{$S$-matcher for $B$ to $D$} if
either $B$ is empty or else $\strat(b_i\gamma) = S$ for some $1 \le i \le k$ and pivot $b_i$.
Let $D_{\preceq S} = \{ a \in D \mid \strat(a) \preceq S\}$, analogously for $\prec$.
For sets $G$ of normal rules let
$G_{\preceq S} = \{ a \CM B \in G \mid \strat(a) \preceq S \tn{ for some $B$}  \}$.
A \define{query} is a normal body. 
\begin{pseudo}*
\hd{Ground}(P,Q,\id{eot}) \ct{Input: $P$ program, $Q$ query, $\id{eot} \ge 0$ ``end of time''.
Output: normal program }\\
$D  \ce \emptyset$ \ct{current domain} \\
$G  \ce \emptyset$ \ct{result ground program} \\
$R  \ce Q$ \ct{current regressed query} \\
for $n \gets 0\..\id{eot}$  do \\+
for $s \gets s_1 \.. s_m$ do \ct{$s_1 \prec \cdots \prec s_m$ is linearization of $P$'s stratification}\\+
  let $S = (n,s)$ \ct{current timed stratum}\\
  let $D^\neg = D_{\prec S}$ \ct{finished domain for strictly earlier timed strata}\\
  for $H \CM B \in P$ do \\+
  for $\gamma$ sth \tn{$\gamma$ is an $S$-matcher of $B$ to $D_{\preceq S}$} do \\+
  for $B' \in \gnd_{D^\neg}(B\gamma)$ do \\+
  if \tn{$B'$ is consistent with $R$} then \ct{see below, assume ``true'' for now}\\+
   let $G' = \pr{Normal}(H\gamma \CM B')$\\
   $D \ce D \cup \{ a \mid \tn{$a \CM B'' \in G'$ for some $B''$ or  $\PR \CC a \in G'$ for some $\PR$}\} $ \\
   $G \ce G \cup G'$\\----
  $R \ce \reg_{\{r \in G_{\preceq S} \mid \tn{$r$ is not a fact}\}}(R)$ \ct{update regressed goal --
    ignore lines 15-17 for now} \\
  $G \ce G \setminus \{ H \CM B \in G \mid \tn{$B$ is inconsistent with $R$} \}$ \\
   $D \ce D \cap \{ a \mid \tn{$a \CM B \in G$ for some $B$ or  $\PR \CC a \in G$ for some $\PR$}\} $ \\-
return $G_{\preceq (\id{eot}, s_m)}$
\end{pseudo}

The \pr{Ground} procedure iterates  over all timed strata, in increasing order, capped
after \id{eot}. It thereby monotonically grows a domain $D$ and a  set of normal rules $G$
obtained by $S$-matchers of the bodies of all rules to
$D_{\preceq S}$, thereby grounding out negative body elements over $D^\neg = D_{\prec S}$.  Notice that while $D$ and $D_{\preceq S}$
can grow while $S$ is fixed, the subset $D^\neg$ remains unchanged. This holds thanks to
using pivot literals stratified
at $S$  in $S$-matchers. For, by SBTP, all ordinary atoms in a
negative body element are lower than $S$, but head atoms are at $S$ or higher and, hence, cannot
contribute to $D^\neg$. For now, the consistency test on line 11 is ignored, i.e., treated
as always  ``true'', and the code on lines 15-17 is igrored. 
See Example~\ref{ex:grounding} for an example for grounding. 



We are going to define the Distribution semantics for \Fusemate programs in terms of their
grounding via the \pr{Ground} procedure and standard (ground level) stratification. The latter enables a
standard fixpoint construction, which includes the standard Distribution semantics for
definite programs as a special case.

\begin{proposition}
\label{prop:SBPT}
  Let $P$ be a program and $n \ge 0$. Then $\pr{Ground}(P,n)$ is SBPT.
\end{proposition}

Let $P_g = \pr{Ground}(P,n)$. From Proposition~\ref{prop:SBPT} it follows that
$P_g$ is stratified in the standard way. For, if $P_g$ were not stratified
then the call graph of $P_g$ has a cycle going through a negative body literals.
This cycle can be lifted to a cycle in the call graph of $P$ by following the 
rules in $P$ corresponding to the ones witnessing the cycle in $P_g$. However, with
Proposition~\ref{prop:SBPT} this cycle does not exist.

This allows us to replace SBTP by standard stratification for ground programs.
As said in the introduction, we do not consider programs with positive cycles here.\footnote{A program like
  \fm{
  fib(0, 0).
fib(1, 1) .
fib(N+1, N1+N2) :- fib(N, N1), N > 0, N <= 20, fib(N-1, N2).
} is in scope as it is SBTP and its grounding has no positive cycle. A symmetry rule
\fm{r(X,Y) :- r(Y,X).} is out of scope.} The call graph of $P_g$, hence, has no cycles.
Similarly to above, let $\prec_g$ denote both the edge relation and 
its transitive closure.
We can now define the distribution semantics of $P$ in terms of least fixpoint models
of probabilistic choices over $P_g$, as follows.


\begin{definition}[Distribution semantics]
\label{def:distribution-semantics}
  Let $P$ be a program, $n \ge 0$ and $P_g = \pr{Ground}(P,n)$. Let $P_g = F \uplus R$ where
$F = \{  \PR \CC a \in P_g \mid \PR < 1 \}$ are all probabilistic facts in $P_g$.
A \define{(probabilistic) choice} is any subset $X \subseteq F$.
Define the probability of the choice $X$ as $\Prob(X) = \Pi\, \{ \PR \mid \PR \CC a \in X\text{ for some $a$}\}  \cdot \Pi\, \{ 1-\PR
\mid \PR \CC a \in F\setminus X\text{ for some $a$}\} $. 
Define \define{$I_R(X)$} as the least fixpoint model of the
(non-probabilistic, $\prec_g$-stratified, normal) program $\{ a \mid \PR \CC a \in X \text{ for
  some $\PR$}\}\cup R$.  Let $Q$ be a query (i.e., a ground normal body).
Define the \define{success probability of $Q$ in $P$} as
$\define{$\Prob(Q\mid P$)} \ce \Sigma\, \{ \Prob(X) \mid X \subseteq F \text{ and } I_R(X) \models Q  \}$.
\end{definition}
In Definition~\ref{def:distribution-semantics},
any $I_R(X)$ is called a \define{model of $P$}, and any  $I_R(X)$ such that $I_R(X) \models Q$ is called a \define{model
  of $Q$ and $P$}. 
An interpretation \define{$I$ satisfies
  right-uniqueness of $\EQ$} iff there are no terms $s$, $t$ and $u \neq t$ such that $\{s\EQ t,\ s \EQ u\} \subseteq I$.
A program $P$ is \define{admissible} if for every $n\ge 0$, each model of $P$
satisfies right-uniqueness of $\EQ$. Admissibility of models is needed for correctness of
our query-guided grounding (Theorem~\ref{th:query-guided-grounding-correctness}).


\paragraph{Query-guided grounding.}
We complete the description of the \pr{Ground} procedure
by definitions of the consistency test and the regression operator $\reg$.
We say that two normal bodies $B_1$
and $B_2$ are \define{consistent} if there is no atom $a$
such that  $\{ a, \neg a \} \subseteq B_1 \cup B_2$ 
and there are no terms $\TT$, $s$, $t$ and $u \neq t$ such that $\{ s \EQ t \AT\TT, s \EQ u \AT\TT \} \subseteq B_1 \cup B_2$.
The term \define{inconsistent} means ``not consistent''.
The idea is that $B_1$ and $B_2$ should be simultaneously satisfiable which is impossible
if one of the these cases applies.\footnote{While in first-order logic $s \EQ t$ and
$s \EQ u$ enforces $t = u$ this is not consistent with the unique name assumptions that is
usually assumed with \emph{Herbrand} interpretations.}
For example, $\{\text{\fm{f(a) = b}}, \text{\fm{f(a) = c}}\}$ is inconsistent. 
See Section~\ref{sec:introduction} for an additional example.
The semantics of consistency is given by right-uniqueness of $\EQ$, as follows:
\begin{lemma}
\label{lemma:inconsistency}
  Let $B_1$ and $B_2$ be inconsistent normal bodies. For every interpretation $I$
  satisfying right-uniqueness of $\EQ$, not both $I \models B_1$ and $I \models B_2$. 
\end{lemma}
\paragraph{Goal regression.} Let $Q$ be a query, $G$ a set of
normal rules, and $a$ an  ordinary atom. (Goal) regression repeatedly expands subgoals $a$ by SLD resolution but keeps only 
new subgoals common to \emph{all}  bodies of the rules for $a$.  Formally,
define \define{one-step regression} $\reg^1_P(a) \ce \bigcap \{ B \mid a \CM B \in P \}$ and
$\reg^1_P(Q) \ce Q \cup \bigcup_{a \in  Q} \reg^1_P(a)$. Define $\reg_P(Q)$ as the
least fixpoint of $\reg^1_P$ starting from $Q$, i.e., $\reg_P(Q) \ce
(\reg^1_P(Q))^k$ where $k$ is the least integer such that $(\reg^1_P(Q))^k = (\reg^1_P(Q))^{k+1}$.
See 
Section~\ref{sec:introduction} for an example.

By \define{query-guided grounding} we mean the \pr{Ground} procedure with
the consistency test on line 11  and the code on lines 15-17 enabled.
By \define{unguided grounding} we mean the condition on line 11
replaced by ``true'' and the code on lines 15-17 disabled.  The following theorem is our main theoretical result.

\begin{theorem}[Correctness of query-guided grounding]
\label{th:query-guided-grounding-correctness}
  Let $P$ be an admissible program and $Q$ be a ground query. Then the success probability of $Q$ in
  $P$ is the same with query-guided grounding and with unguided grounding.
\end{theorem}


\section{Inference}
\label{sec:inference}
We implemented the grounding algorithm of Section~\ref{sec:grounding} in our \Fusemate
system. We also implemented functionality for probabilistic query answering for
non-ground conditional queries based on a ground-level variable
elimination algorithm at its core. In this section we briefly describe the latter components.

An \define{conditional input query} has the form $B \mid E$ consisting
of a body $B = \QM b_1,\ldots,b_k, \NOT \vec c_1, \ldots ,\NOT \vec c_n$
as in \eqref{eq:rule} and optional \define{evidence} $E = e_1,\ldots,e_m$ of ground
ordinary atoms (see Section~\ref{sec:introduction} for examples).  
Given $B \mid E$ and a program $P$, we want to compute the success probability of $B \mid E$
in $P$.
We do this by grounding  $B \mid E$ and $P$ and reduction to query success
probability (Definition~\ref{def:distribution-semantics}). 
The grounding executes several calls to \pr{Ground}(P,Q,\id{eot}): (a) a call with
query $Q = E$, (b) a call with query $Q$ comprised of all ground literals in  the
conjunction $(B,E)$, and (c) calls with queries $Q = (B\gamma,E)$,
for every $S$-matcher for $B$ to domain $D$ computed in \pr{Ground} in step (b) and
biggest stratum $S$. The value for \id{eot} is given as input.

The rationale for this staged process is to obtain in (b) a domain $D$ as small as
possible for exhaustive grounding of $(B,E)$ in (c). 
The result reported to the user are
answer substitutions $\gamma$ and their probabilities, i.e., pairs $(\gamma, \Prob(B\gamma\mid E))$ where $\Prob(B\gamma\mid E) = \Prob(B\gamma,E\mid P) / \Prob(E \mid P)$ for
every \define{answer substitution} $\gamma$ in step (c). Notice that for each of these
computations the optimized ground programs $P_g$  computed by \pr{Ground} in steps
(a), (b) and (c) are available for that. 
Theorem~\ref{th:query-guided-grounding-correctness} guarantees the correctness.

With the above considerations, the algorithmic problem now is to compute $P(Q\mid P_g)$
for a query $Q$ and ground program $P_g$. Many PLP systems would convert $P_g$ into an equivalent
weighted Boolean formula and apply weighted model counting algorithms on its d-DNNF normal
form (or other)~\cite{fierens_inference_2015,chavira_probabilistic_2008}.
For now, \Fusemate implements a variable elimination algorithm based on finding
all SLD proofs of the query.

To explain the main ideas, every SLD proof provides a factor by multiplying the probabilities of the facts at its leaves.
The success probability of $Q$ in $P_g$ then is the sum of the probabilities
of these factors. For correctness, the bodies of all rules with the same head have to be made
disjoint (inconsistent) first. This can be achieved with indicator
variables~\cite{poole_probabilistic_1993}:
every collection $\{h \CM B_1, \cdots, h \CM B_m\}$ of all rules in $P_g$ with  the same head
$h$ is replaced by the  rules
 $\bigcup_i (\{ h_i \CM B_i\} \cup \{ h \CM \neg h_1,\ldots, \neg h_{i-1},h_i\})$
 where $h_i$ is a fresh predicate symbol indicating which case is tried exclusively. 
(Probabilistic facts are not affected, as they are all made with pairwise distinct
atoms, by design of the \pr{Normal} procedure.)
Our variable elimination algorithm \pr{VE} can be applied to such programs and (ground,
conjunctive) queries $Q$. It caches intermediate results, and it 
prunes inconsistent queries as soon as they come up: 
\begin{pseudo}*
  \hd{VE}(P_g, Q) \ct{Input: $P_g$ normal program with disjoint bodies, $Q$ query. Output: $\Prob(Q\mid P)$}\\
  $\id{Cache} := \emptyset$ \ct{cached pairs (Query, Probability)}\\
  \pr{Return}(Q, R) \ct{Caches and returns result $R$ for $Q$} \\+
    $\id{Cache} \ce \id{Cache} \cup \{ (Q, R)\}$\\
  $R$\\-
  \pr{Inner}(Q) \ct{Input: worked off and remaining query literals $Q$. Output:
    probability of $Q$} \\+
  if $Q = \emptyset$ then $1.0$ \\
  else if \tn{$Q$ is inconsistent} then $0.0$ \ct{from here on $Q$ is consistent}\\
  else if $(Q,\PR) \in \id{Cache}$ then $\PR$\\
  else let $L \in Q$ sth \tn{$L$ is maximal in $Q$ wrt.\ the stratification order $\prec_g$}  \\+
 let $Q' = Q \setminus \{L\}$ \\
  if $L = \neg A$ then \ct{$\Prob(\neg A,Q') = \Prob(Q') - \Prob(A,Q')$}\\+
  \pr{Return}(Q, \pr{Inner}(Q') - \pr{Inner}(Q' \cup \{A\}))\\-
  else if $\PR \CC L \in P_g$ then \\+
   \pr{Return}(Q,  \PR \cdot \pr{Inner}(Q'))\\-
  else let $Bs = \{ B \mid L \CM B \in P_g \}$ \ct{$Bs$ is new subgoals as per rules for $L$ in $P_g$}\\+
    \pr{Return}(Q, \Sigma_{B \in Bs} \pr{Inner}(B \cup Q'))\\---
\pr{Inner}(Q)
\end{pseudo}
The main ideas behind \pr{VE} should be clear from the comments above, but some notes are
in order. Starting from \pr{Inner}(Q) the query literals in $Q$ are successively
resolved away using facts and rules from $P_g$ (except negative literals, which are
expanded in the obvious way in line 12).  Line 7 is the mentioned early pruning check.
For example, let $P_g$ contain, among others, \fm{0.5::p. a=1 :- p. a=2 :- -p.} and $Q = \{\ldots,\text{\fm{a=1, a=2}}\}$. Notice that the stated rules do not offend
admissibility as no model can make both \fm{a=1} and \fm{a=2} simultaneously true.
That is,  the conjunction
\fm{a=1, a=2} does not hold in any model. Early pruning 
discovers this fact via inconsistency of  \fm{a=1, a=2} and assigns
probability 0 to $Q$. This avoids redundantly solving potentially present subgoals of $Q$ that are greater wrt. $\prec_g$
than \fm{a=1} and \fm{a=2}.
Another points to notice is that no probabilistic fact is multiplied in more than once. This
follows from working off the query literals in decreasing stratification order.






\section{Application Examples and Benchmarks}
\label{sec:application-examples}
In this  section we evaluate the practical effectiveness of \Fusemate and query-guided grounding. 
\Fusemate and the problem sets described below can be downloaded
from \url{https://bitbucket.csiro.au/projects/MOVEMENTANALYTICS/repos/fusemate-distrib/}.  
All experiments were run on a Dell Latitude 7330 laptop, with Windows 11 operating system
(32 GB RAM, 3.2 GHz).
For each problem, we compared timings in \Fusemate and ProbLog by running several sets of
queries for increasing complexity in the query. We only summarize our findings
here. More details on the ProbLog and \Fusemate encodings are in the Appendix.



\begin{figure}[h]
\begin{minipage}[b]{5cm}
\begin{minted}[fontsize=\footnotesize]{prolog}
%% Markov Model
in ~ [a, b, c] @ 0.
in ~ [[a, 0.9],[b, 0.05],
          [c, 0.05]] @ T+1 :- in=a @ T.
in ~ [[a, 0.7],[c, 0.3]] @ T+1 :- in=b @ T.
in ~ [[a, 0.8],[c, 0.2]] @ T+1 :- in=c @ T.

%% Time steps N = 20
?- in=a@0, in=a@1,.., in=a@20.

%% Specificity, N = 7
?- in=a@0, in=a@1, in=L2@2,..,in=L8 @ 8.

%% Timepoint, N = 20
?- in=a@23.

\end{minted}
\end{minipage} \hspace*{3em}
  \includegraphics[scale=0.67]{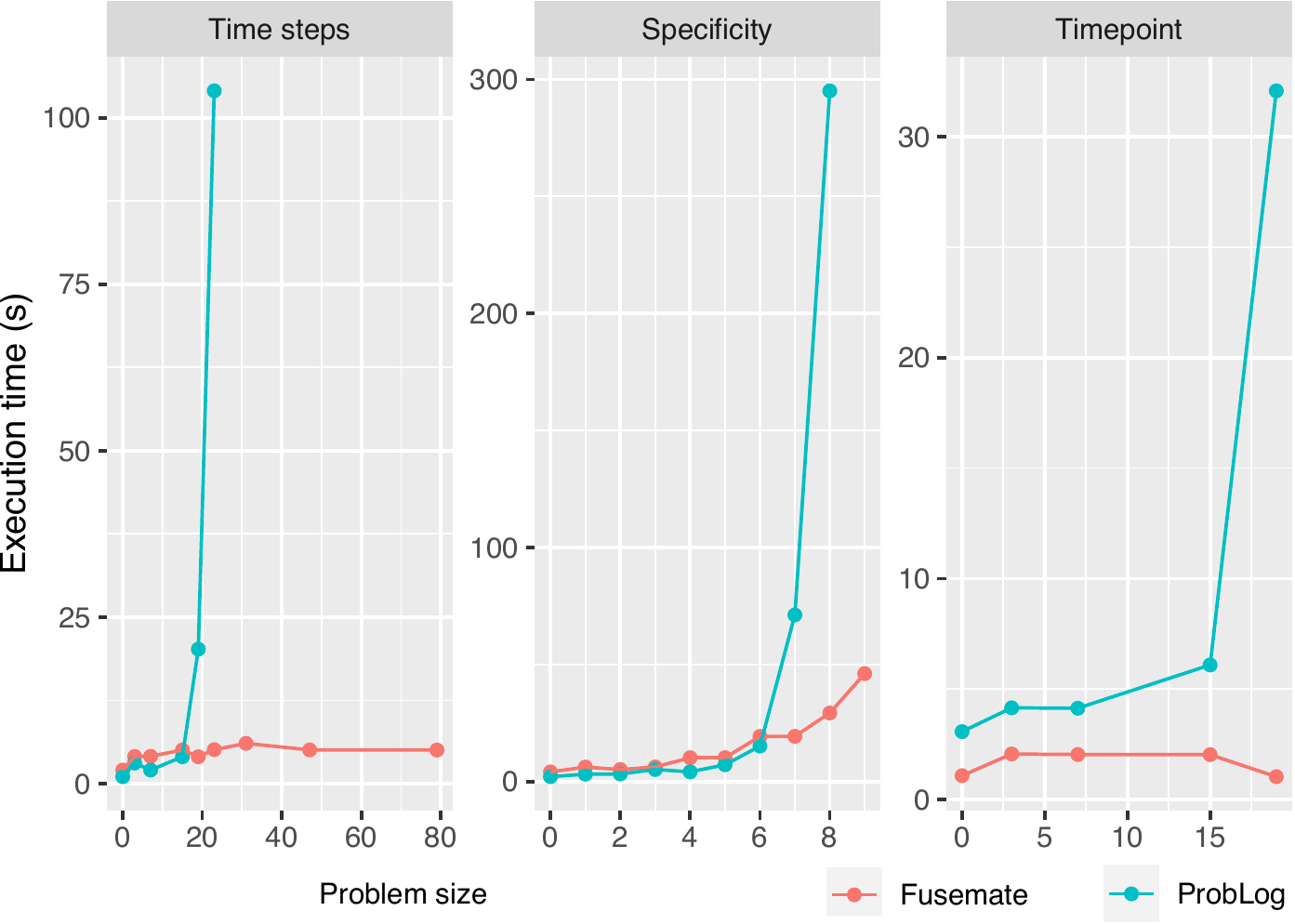}

\caption{Timings for three kinds of queries to the Markov chain in \Fusemate (red)
  and ProbLog (blue). }
\label{fig:markov-chain}
\end{figure}

\paragraph{Markov model.} 
This problem is from the ProbLog tutorial Markov chain
example.~\footnote{\url{https://dtai.cs.kuleuven.be/problog/tutorial/various/08_bayesian_dataflow.html}}
The code models the movement in time 
between three locations $a$, $b$ and $c$, where the probability of moving to the next location depends only
on the current location.
We timed three kinds of queries which we refer to as \emph{timesteps}, \emph{specificity} and \emph{timepoint}.
The timesteps query asks for the probability of being at position $a$ during a
whole period of $N=0,\ldots,80$ timesteps.
The specificity query asks for the probability of being in a certain location across nine
timesteps, where complexity $N$ is increased by decreasing the specificity of those
locations. We do that by replacing fixed location $a$ with a variable, so the program has
to repeat the calculation for all possible locations, increasing the branching.  
The timepoint query asks for the probability of being at location $a$ at a final time
point $N$.
The logic program, sample queries and runtime performance are in Figure~\ref{fig:markov-chain}.

Results in
Figure \ref{fig:markov-chain} indicate that \Fusemate performs better than ProbLog for
increasing problem complexity. The \Fusemate times are measured with query-guided grounding on.
However, \Fusemate can also solve the problems without it. Runtimes and number of ground rules generated raise by
a factor of 2-3 without guidance.
It seems that \Fusemate's variable elimination procedure is better suited to this example
than the ProbLog inference engine. Inconsistency pruning (line 7 in \pr{VE}) did not have
an impact on these problems.

\paragraph{Hidden Markov model.} 
This example is based on the example from the Wikipedia page for hidden
Markov models\footnote{\url{https://en.wikipedia.org/wiki/Hidden_Markov_model}}.
It models the situation where a prediction of the weather
(sunny/rainy) is determined by observations of how much water due to precipitation has
been collected in a bowl that is placed outside.  The idea is that more water will be collected when it rains more, which is the measurement of how the weather has been. For simplicity, this example assumes no evaporation, so the observations of water in the bowl are non-decreasing. The logic program was described in Section~\ref{sec:introduction}.

\begin{figure}[h]
\begin{minipage}[b]{5.5cm}
\begin{minted}[fontsize=\footnotesize]{prolog}
%% See Introduction for program

%% Queries for N=3
%% Sunny
?-state=X @ 3 | obs=0 @ 1, obs=0 @ 2, obs=0 @ 3.

%% Rainy
?-state=X @ 3 | obs=4 @ 1, obs=8 @ 2,  obs=12 @ 3.

%% Mixed
state=X @ 3 | obs=0 @ 1, obs=4 @ 2, obs=24 @ 3.

\end{minted}
\end{minipage} \hspace*{3em}
\includegraphics[scale=0.67]{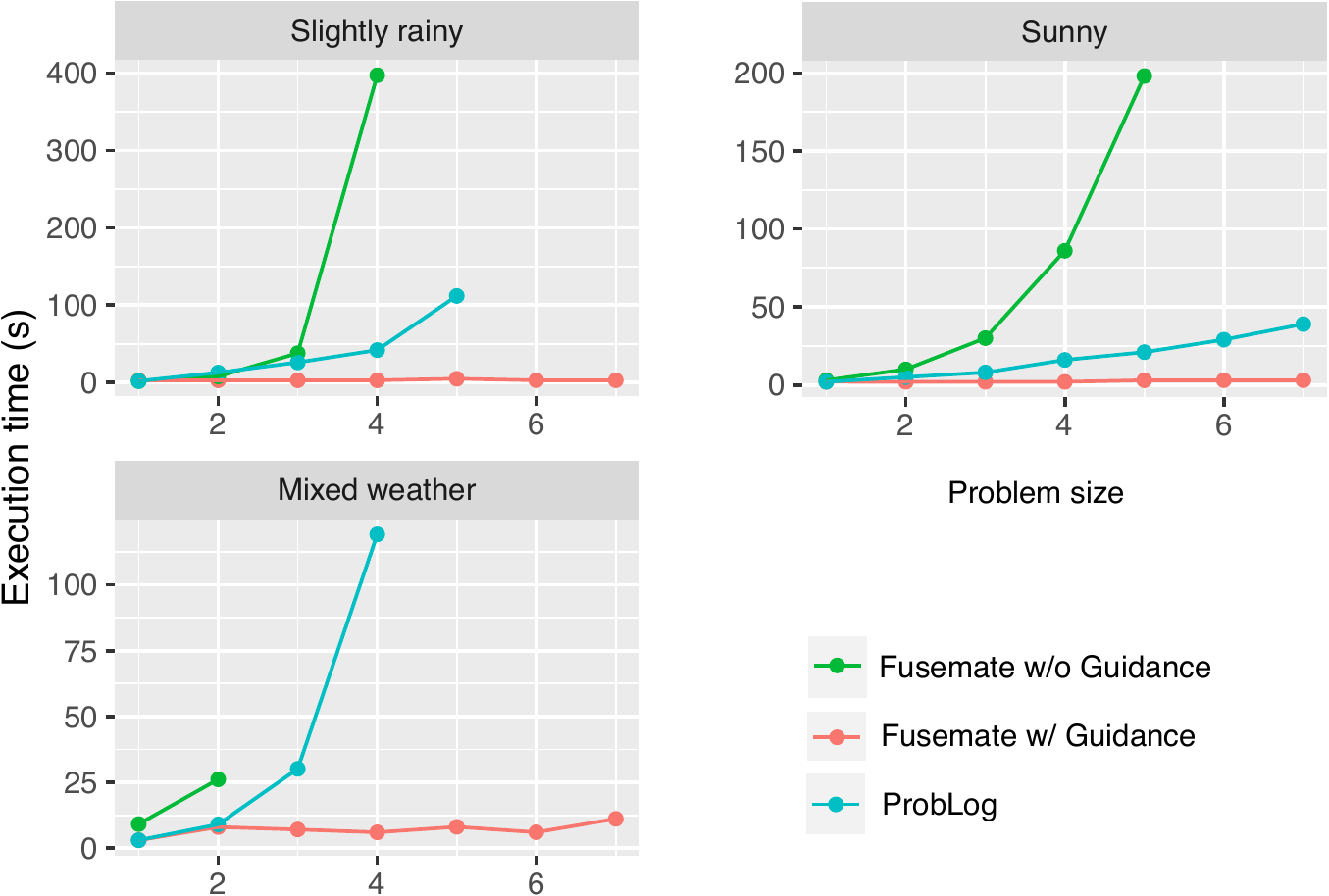}

\caption{Timings for three kinds of queries to the hidden Markov model precipitation example:
\Fusemate without query-guidance (green), \Fusemate with
  query-guidance (red), and ProbLog (blue). }
\label{fig:precipitation}
\end{figure}

We timed three kinds of filtering queries for estimating the current state based on past
observations trending around \emph{slightly rainy}, \emph{sunny}, and \emph{mixed},  scenarios, respectively.
In the slightly rainy scenario each timepoint increased the amount of water observed by
four, a value that can occur from either the sunny or the rainy distributions. 
In the sunny scenario, each observation was of zero
precipitation, so the observations must have come from the sunny distribution.
In the mixed weather scenario, a variety of observations were included in this query,
which meant there was a mixture of observations from both the sunny and rainy
distributions. For each scenario, the query complexity $N$ was increased by increasing the
number of observations and the timepoint at which the final state is predicted. 
The results and sample queries are described in Figure \ref{fig:precipitation}.


For all problems, \Fusemate with unguided grounding was the slowest, but \Fusemate with
query-guided grounding consistently outperformed ProbLog.
 What sets this example apart from the Markov Chain example is its high branching
rate (30 support values in a rain state, vs.\ two or three Markov Chain) ,
 and the implicit dependence of each state on its history via the accumulated rainfall up
to that  state. 

Inconsistency pruning did not play a role in \Fusemate's
good performance in the problem above. We tested this by disabling the test on line 7 in the \pr{VE} procedure.
With less constraining queries, however, inconsistency pruning can lead to drastic
performance improvements. 
We experimented with relaxing the evidence of a slightly modifed ``sunny'' scenario by
leaving out observations. For a query size $N=4$, for instance, we obtained the following runtime results
(in seconds):

{\centering\small
\qquad\qquad\begin{tabular}{lrr}
& \multicolumn{2}{c}{\bfseries Inconsistency pruning} \\
\bfseries Query & \bfseries Off & \bfseries On \\\hline
\fm{?- state=S @ 4 | obs=0 @ 1, obs=0 @ 2, obs=0 @ 3, obs=10 @ 4} & 7.5 & 7.5 \\
\fm{?- state=S @ 4 | obs=0 @ 1, obs=0 @ 2,                     obs=10 @ 4} & 44.5 & \bfseries 13.5 \\
\fm{?- state=S @ 4 | obs=0 @ 1,                                          obs=10 @ 4} & >2000 & \bfseries 30.0 \\
\fm{?- state=S @ 4 |                                                              obs=10 @ 4} & >2000 & \bfseries 180.75\\\hline
\end{tabular}
}

In addition to overall solution times we were also interested in comparing groundings. 
For ProbLog we observed unusual high grounding times, and for Fusemate with
unguided grounding we observed large but quickly computed groundings (which become
unmanageable for inference quickly). For the mixed weather problem, we observed:

{\centering\small
\qquad\qquad\begin{tabular}{r@{\qquad}rrcrrr}
& \multicolumn{2}{c}{\bfseries\Fusemate \#ground rules} & \multicolumn{3}{c}{\bfseries ProbLog}\\
  $N$ & \multicolumn{1}{c}{\bfseries query-guided} & \multicolumn{1}{c}{\bfseries
                                                     unguided} & \quad &
                                                                 \multicolumn{1}{c}{\bfseries total time} & \multicolumn{1}{c}{\bfseries grounding time} & \multicolumn{1}{c}{\bfseries \#ground rules} \\\hline
2 &  2200 &  6500 & & 9.0 &  8.3 &  53 \\
3 &  2270 &  12900 &  & 30 &  19 & 276 \\
4 &  2300 &  21400 &  & 119 &  33 & 499 \\
5 & 2400 & 32000 & & & 50 & 682\\
6 & 2470 & 45000 & & & 65 & 839\\
7 & 2500 & 60000 & & & 95 & 1068\\\hline
\end{tabular}
}

Grounding sizes between \Fusemate and ProbLog are only roughly comparable. \Fusemate
outputs ground normal rules,  where ProbLog outputs ground rules with annotated
disjunctions, where head probabilities are left in place. The \Fusemate total times are between 1 and 7 seconds and not listed in
the table. For ProbLog, grounding time is still well below solving time but well above
\Fusemate total times. 

%



\section{Conclusion}
In this paper, we proposed a bottom-up grounding approach for an expressive
probabilistic logic programming language (expressive form of stratification, expressive
default negation, dynamic distributions).
We defined the semantics of the input languages as an extension of the standard Distribution semantics via a
standard fixpoint construction after grounding and transforming away default
negation. 
As the main contribution of this paper, we integrated and proved correct a novel
technique for avoiding ground instances that are irrelevant for proving a given query.
Grounding, transforming away default negation, and query-guided pruning are
tightly integrated and carried out incrementally along the program's stratification
order. They rest on a built-in semantics for equations as right-unique relations, which is
appropriate, e.g., for representing (finite) distributions. 

We showed the effectiveness of query-guided pruning experimentally.
The rationale is to tackle combinatorial explosion \emph{during} grounding instead
of attempting optimizations \emph{afterwards}.
Without guidance, example problems tend to grow to unmanageable size quite quickly.
Our system outperformed ProbLog on hidden Markov model filtering problems with a high branching rate.
(On other domains not reported here  we found that ProbLog often performs 
better than \Fusemate.) We also showed that the performance of our top-down variable elimination algorithm 
benefits from building-in inconsistency pruning.
While the result are somewhat limited in scope, we 
suggest that the research direction begun in this paper looks promising for further exploration.
We will consider optimized off-the-shelf  backends for weighted
inference as a possibly better alternative to our variable elimination algorithm in such cases.
We also plan to integrate a magic set transformation, which is complementary to our query-guided grounding.

We conjecture that query-guided grounding enables \Fusemate to solve filtering
queries in linear time. This would be the same complexity as the dedicated
forward-backward algorithm. The proof hinges on an analysis of the
solution caching mechanism in \Fusemate's variable
elimination inference algorithm. 

\paragraph{Acknowledgements.} We thank the reviewers for their valuable comments. 





{\fontsize{10pt}{12pt}\selectfont

}


\clearpage
\newpage
\appendix

\section{Proofs}
\label{sec:proofs}

\begin{lemma*}[Grounding preserves semantics]{lemma:grounding}
Let $D$ a domain, $I \subseteq D$ and $\vec c$ a sequence of atoms.
Let $E = \{ \vec c\gamma_1, \ldots,\vec c\gamma_k \}$ where $\gamma_1,\ldots,\gamma_k$ are all matchers of
$\vec c$ to $D$ ($k=0$ is possible) and $H$ the set of all hitting sets of $E$.
Then $I \models \NOT \vec c$ iff there is a $\{d_1,...,d_k\} \in H$ such that $I \models \neg d_1,...,\neg d_k$.
\end{lemma*}
\begin{proof}
(``$\Rightarrow$'') Assume $I \models \NOT \vec c$. 
None of the $\gamma_1,\ldots,\gamma_k$ is a matcher of $\vec c$
to $I$ because it would contradict the assumption $I \models \NOT \vec c$. 
In other words $I \not\models \vec c\gamma_i$, for all $1 \le i \le k$. Equivalently, for every $1 \le i \le k$ there is a $d_i \in \vec
c\gamma_i$ such that $I \not\models d_i$. By definition of hitting set, $\{ d_1,\ldots, d_k\} \in H$, and
from $I \not\models d_i$ it follows $I \models \neg d_1,\ldots,\neg d_k$. 

(``$\Leftarrow$'') By contradiction. Assume there is a set $\{d_1,...,d_k\} \in H$ such that $I \models \neg
d_1,...,\neg d_k$ but there is a matcher $\gamma$ for $\vec c$ to $I$. 
By construction, $\{d_1,...,d_k\}$ is a hitting set of $E = \{ \vec c\gamma_1, \ldots,\vec c\gamma_k \}$.
Because $I \subseteq D$ the assumed matcher $\gamma$ of $\vec c$ to $I$ is also a matcher of
$\vec c$ to $D$. (Just select in $D$ the same ordinary atoms as in $I$.) That is, $\gamma
= \gamma_i$ for some $i \le i \le k$ (because $\gamma_1,\ldots,\gamma_k$ are \emph{all}  matchers).
As a  matcher, $I \models \vec c\gamma$. Hence, trivially $I \models \vec c\gamma_i$. 
From the assumption $I \models \neg d_1,...,\neg d_k$ it follows $I \not\models d_i$. By hitting set $d_i \in \vec
c\gamma_i$. Together then $I \not\models \vec c\gamma_i$, a plain contradiction.
\qed
\end{proof}

\begin{proposition*}{prop:SBPT}
  Let $P$ be a program and $n \ge 0$. Then $\pr{Ground}(P,n)$ is SBPT.
\end{proposition*}
\begin{proof} (Sketch)
  By definition, $P$ is SBTP. Applying a matcher $\gamma$ to a rule $H \CM B \in P$ trivially preserves SBTP.
  Grounding the body $B\gamma$ (Proc.~\ref{proc:grounding}) preserves SBTP because
  every element in the domain
  $D_{\prec S}$ for that is $\prec$ than $S$, and, by definition of $S$-matcher there is a pivot in $b \in B$ with
  $\strat(b\gamma)  = S$. It is not difficult to check that the auxiliary definitions introduced
  by \pr{Normal} are a conservative extension of $P$'s SBPT.
  \qed
\end{proof}

\begin{lemma*}{lemma:inconsistency}
  Let $B_1$ and $B_2$ be inconsistent normal bodies. For every interpretation $I$
  satisfying right-uniqueness of $\EQ$, not both $I \models B_1$ and $I \models B_2$. 
\end{lemma*}
\begin{proof}
Assume an $I$ as stated such that $I\models B_1$. We show $I\not\models B_2$ directly.
By consistency, in the first case $\{ a, \neg a \} \subseteq B_1 \cup B_2$
for some $a$. Because $I \models B_1$ it follows trivially $\{ a, \neg a  \} \not\subseteq B_1$. If $\{ a, \neg a
\} \subseteq B_2$ then $I \not\models B_2$ follows trivially. Otherwise, either $a \in B_1$ and $\neg a \in B_2$ or
$\neg a \in B_1$ and $a \in B_2$. In the first case, $I \models B$  entails $I \not\models \neg a$ and hence $I \not\models 
B_2$. In the second case, $I \models B$ entails $I \not\models a$ and hence $I \not\models \neg B_2$, too. This covers
all cases.

By consistency, in the second case $\{ s \EQ t \AT\TT, s \EQ u \AT\TT \} \subseteq B_1 \cup B_2$ for
some terms $s$, $t$ and $u \neq t$. With right-uniqueness of $I$ the proof is literally the
same as in the first case after substituting $a$ by $s \EQ t \AT\TT$ and $\neg a$ by $s \EQ u \AT\TT$.
  \qed
\end{proof}

\begin{theorem*}[Correctness of query-guided grounding]{th:query-guided-grounding-correctness}
  Let $P$ be an admissible program and $Q$ be a ground query. Then the success probability of $Q$ in
  $P$ is the same with query-guided grounding and with unguided grounding.
\end{theorem*}
\begin{proof} Referring to Definition.~\ref{def:distribution-semantics}, 
  let $n \ge 0$ and $P_g$  ($P_g^\q$) be the program obtained from unguided (query-guided) grounding, respectively.
By an ``unguided model'' we mean any model $I$ of $Q$ and $P$ obtained from $P_g$ according to
Definition~\ref{def:distribution-semantics}. 
Analogously, a ``guided model'' $I_g^\q$ is a model of $Q$ and $P_g^\q$.
We show that every guided model is also an unguided model, and vice
versa. This suffices to prove the theorem because every other $I$ with $I \not\models Q$
does not contribute to the success probability of $Q$ in $P$, and analogously for $I^\q$.
In the first part of the proof we ignore query regression but  return to it afterwards.

For the first direction, let $I$ be an unguided model. We have to show that it is also a
guided  model $I^\q$. The proof is by induction on the ordering $\prec_g$ used for the least
fixpoint construction of $I$. The induction start with $I = \emptyset = I^\q$ is trivial.
The induction step is the transition $I \to I \cup \{a\}$ where there is a rule $a \CM B'' \in P_g$ such
that $I\models B''$ and $I \not\models a$. (This is by definition of lfp construction.)
As the induction  hypothesis suppose that $I$ is a guided  model $I^\q = I$.
We have to show that $I^\q \cup\{a\}$ is also a guided  model.

In the unguided  model a rule $a \CM B''$ must have been obtained starting with an
$S$-matcher $\gamma$ for $B$ to $D_{\preceq S}$ for some timed stratum $S$, domain $D$ and rule
$H \CM B \in P$ constructed in the unguided run of \pr{Ground}(P,Q,n), see lines 9-14.  Let
$B'$ be the chosen ground body on line 10.

The \pr{Normal} transformation does
not affect the truth value of $B'$ when transformed to $B''$. For that we need that the
newly introduced atoms are suitable stratified so that their truth value is fixed in $I$
and not changed later. This can be checked by inspection of  the definition of \pr{Normal}.
With $I \models B''$ it follows $I \models B'$.

By Lemma~\ref{lemma:grounding} and SBTP,
grounding of $B\gamma$  on line 10 does not affect the truth value in
$I$.  For that we need that fact that $I \subseteq D$. This follows from the lfp construction by which
every $a \in I$ must stem from a probabilistic fact $\PR\CC a$ or be the head of a normal rule from
$G$. These are all in $D$ according to lines 12 and 13 in \pr{Ground}.
With $I \models B'$ it follows $I \models B\gamma$.

We inspect the run of \pr{Ground}(P,Q,n) for the guided  model. The interpretation $I^\q$
is a subset of a set $D$ constructed at some point in this run, for the same reason as
above.
(As a remark, while by induction
hypothesis $I^\q  = I$, the set $D$ in the guided run can be a subset of the set $D$ in the unguided one.)
It follows that the same $S$-matcher $\gamma$ for $B$ to $D_{\preceq S}$ as above exists.
That is, the for-loop on line 9 will select this $\gamma$ and reach line 10.

From $I^\q = I$  it follows trivially $I^\q \models B\gamma$.
As in the unguided case, grounding does not affect satisfaction.
More precisely, by Lemma~\ref{lemma:grounding} and SBTP, the for-loop on line 10 will select a
$B'$ such that $I^q \models B'$.
By guided model property $I^\q \models Q$. With $I^\q \models B'$ 
and Lemma~\ref{lemma:inconsistency} in the contrapositive
direction it follows that $B'$ and $Q$ are consistent.
That is, line 12 will be reached and $G'$ will contain the rule $a \CM B''$, for some $B''$.
The bode $B''$ represents the same probabilistic choice as the
corresponding rule in the unguided case (also named $a \CM B''$ above) which is satisfied
by $I$. It follows that $I^\q$ satisfies the same choice. That is, with $I^\q\models B'$  it
follows $I^q \models B''$. As  $a \CM B'' \in P_g^\q$ the lfp construction will  apply this rule and
derive $I^\q \cup \{a\}$ as a model of $P_g^\q$.
Trivially, with $I \cup \{a\}$ being a model of $Q$ and $P$ so is $I^\q \cup \{a\}$ as desired.

For the other direction, let $I^\q$ be a guided model. We have to show that it is also an
unguided model $I$. Like in the first case the proof is by induction and has the
same shape. We do not spell it out in detail. The main argument lies in the fact that in unguided grounding the
condition in line 11 is always ``true''. More precisely, every rule in $P_g^\q$ has a
counterpart in $P_g$ with the same head but possibly different body stemming from a
grounding over a different set $D$. As in the first case, this makes no difference regarding
satisfaction of the bodies in $I$ and $I^\q$, respectively.
In conclusion, this makes $I\cup \{a\}$ a model of $P$, and
with $I \cup \{a\}$ being a model of $Q$ and $P$ so is $I^\q \cup \{a\}$ as desired.

Finally, turning to regression. The current query $R$ is initialized with $Q$ but strengthened
in the course of the run. In its core, we need to show why adding literals to $R$ does not
eliminate models of $Q$ and $P$, hence preserves success probability of $Q$ in $P$.
(Trivially, making the query stronger cannot enable
additional models.)

In the following assume a model $I^\q$ of $Q$ and $P$.
The proof is by induction along the timed stratification order. The
induction start with $R = Q$ is trivial, as no literals are added.
For the induction step assume $I^\q \models R$ and that $R \ce R \cup Q'$ on
line 15 for some set $Q$ of literals obtained be regression with the rules $M$.
We have to show $I^\q \models R \cup Q'$

The rules in $M$ all have a head with stratum $S$, and no more rules with head stratum $S$ will be
computed later in the run of \pr{Ground}. (The computation continues with a higher stratum
from then on.)
We assumed $I^\q \models R$. This means that every atom $a \in R$ is either a fact in $I^\q$ or contained in the
least fixpoint model as per Definition~\ref{def:distribution-semantics}. In the latter
case it follows there is a rule $a \CM B \in P_g^\q$ such that $I^\q \models B$. This rule must be from the set $M$.

By construction, regression extends $R$ with literals that are common to all bodies of rules in
$M$ with the same head, in this case, $a$. Let $B' \subseteq B$ be that (possibly empty) set of
common literals. As $I^\q \models B$
trivially entails $I^\q \models B'$ the case for $a$ is done and it follows $I^\q \models R \cup B'$. The same argument is
repeated for all other extensions until the fixpoint $R \cup Q'$ of extensions is reached.
Formally, this is by itself an inductive proof and yields $I^\q \models R \cup Q'$ as required.

Some comments on the code on lines 15 and 16. The code on line 15 removes ground
rules from $G$ whose body is inconsistent with $R$. No such removed rule can contribute to
a model of $R$ and $P$ because of inconsistency. The code on line 16 removes a domain
element if it was contributed earlier by such a removed rule only. 
\qed
\end{proof}

\clearpage\newpage

\section{Code for Examples and Benchmarks}
\label{app:code-examples-benchmarks}

In this appendix, we give the full details and code for the examples and benchmarks in Section \ref{sec:application-examples}. Each example was tested on a set of queries. Within each set, the complexity $N$ increases, so we can compare the runtimes of \Fusemate and ProbLog as they tackle problems that require more computation.

\subsection{Markov model}
\label{app:markov-chain}

This example is based on a ProbLog
tutorial\footnote{\url{https://dtai.cs.kuleuven.be/problog/tutorial/various/08_bayesian_dataflow.html}} and describes movement between three locations: $a$, $b$ and $c$. It is a Markov model because the probability of moving to the next location depends only on the current location.

The model is defined in \Fusemate as follows

\begin{minted}{prolog}
% initial distribution
in ~ [a, b, c] @ 0.
% transition matrix and define walk
in ~ [[a, 0.9], [b, 0.05], [c, 0.05]] @ T+1 :- in=a @ T.
in ~ [[a, 0.7], [c, 0.3]] @ T+1 :- in=b @ T.
in ~ [[a, 0.8], [c, 0.2]] @ T+1 :- in=c @ T.
\end{minted}

The equivalent ProbLog code is
\begin{minted}{prolog}
% transition matrix
0.9::go(a,a,T); 0.05::go(a,b,T); 0.05::go(a,c,T).
0.7::go(b,a,T); 0.3::go(b,c,T).
0.8::go(c,a,T) ; 0.2::go(c,c,T).

% add initial distribution and define walk
1/3::in(a,0); 1/3::in(b,0); 1/3::in(c,0).
in(X,T) :- T > 0, TT is T-1, in(Y,TT), go(Y,X,TT).
\end{minted}

We timed three sets of queries which we refer to as timesteps, specificity and timepoint.

\paragraph{Timesteps.} The first calculates the probability of being at position $a$ at every time step for $N=0,\ldots,80$ timesteps. For example, in \Fusemate at $N=0$ the query is
\begin{minted}{prolog}
% timesteps
config(eot, 0).
?- 
   in=a @ 0.
\end{minted}
and at $N=4$ the query is
\begin{minted}{prolog}
% timesteps
config(eot, 4).
?- 
    in=a @ 0,
    in=a @ 1,
    in=a @ 2,
    in=a @ 3,
    in=a @ 4.
\end{minted}

We ran the equivalent code in ProbLog, but stopped at $N=19$.
\begin{minted}{prolog}
% timesteps: N=0
q(X0) :-
    in(X0,0).
query(q(a)).
\end{minted}
\begin{minted}{prolog}
% timesteps: N=4
q(X0, X1, X2, X3, X4) :-
    in(X0,0),
    in(X1,1),
    in(X2,2),
    in(X3,3),
    in(X4,4).
query(q(a, a, a, a, a)).
\end{minted}

\paragraph{Specificity.} The second example calculates the probability of being in a certain location across nine timesteps, where complexity $N$ is increased by decreasing the specificity of those locations. We do that by replacing fixed location $a$ with a variable, so the program has to repeat the calculation for all possible locations, increasing the branching. 

In \Fusemate, the $N=0$ case is
\begin{minted}{prolog}
% specificity
config(eot, 8).
?-
    in=a @ 0,
    in=a @ 1,
    in=a @ 2,
    in=a @ 3,
    in=a @ 4,
    in=a @ 5,
    in=a @ 6,
    in=a @ 7,
    in=a @ 8.
\end{minted}

and $N=9$ is
\begin{minted}{prolog}
% specificity
config(eot, 8).
?-
    in=L0 @ 0,
    in=L1 @ 1,
    in=L2 @ 2,
    in=L3 @ 3,
    in=L4 @ 4,
    in=L5 @ 5,
    in=L6 @ 6,
    in=L7 @ 7,
    in=L8 @ 8.
\end{minted}

Equivalently, in ProbLog we define
\begin{minted}{prolog}
 q(X0, X1, X2, X3, X4, X5, X6, X7, X8) :-
     in(X0,0),
     in(X1,1),
     in(X2,2),
     in(X3,3),
     in(X4,4),
     in(X5,5),
     in(X6,6),
     in(X7,7),
     in(X8,8).
\end{minted}
and then for $N=0$ query
\begin{minted}{prolog}
% specificity
query(q(a, a, a, a, a, a, a, a, a)).
\end{minted}
and replace each of the constants with variables until for $N=9$
\begin{minted}{prolog}
% specificity
query(q(L0, L1, L2, L3, L4, L5, L6, L7, L8)).
\end{minted}

\paragraph{Timepoint.} In this final example, we calculate the probability of being at location $a$ at the final timepoint, increasing the value of that timepoint to increase complexity.

In \Fusemate, we start at $N=0$
\begin{minted}{prolog}
% timepoint
config(eot, 0).
?- 
    in=a @ 0.
\end{minted}
and finish at $N=79$
\begin{minted}{prolog}
% timepoint
config(eot, 79).
?- 
    in=a @ 79.
\end{minted}

\subsection{Hidden Markov model (precipitation)}
\label{app:hmm-precipitation}

The situation modeled in this example is that of predicting the weather (rainy/sunny) from observations of how much water due to precipitation has been collected in a bowl that is placed outside. The idea is that more water will be collected when it rains more, which is the measurement of how the weather has been. For simplicity, this example assumes no evaporation, so the observations of water in the bowl are non-decreasing.

The model in \Fusemate is
\begin{minted}{prolog}
config(inst_sol, true).
config(show_info, false).
config(cautious_disjointing, true).
%% start_probability = {'Rainy': 0.6, 'Sunny': 0.4}
state ~ [[rainy, 0.6], [sunny, 0.4]] @ 0.

%% transition_probability = {
%%    'Rainy' : {'Rainy': 0.7, 'Sunny': 0.3},
%%    'Sunny' : {'Rainy': 0.4, 'Sunny': 0.6},
%% }

state ~ [[rainy, 0.7], [sunny, 0.3]] @ T+1 :- state=rainy @ T.
state ~ [[rainy, 0.4], [sunny, 0.6]] @ T+1 :- state=sunny @ T.

%% obs=R @ T : R is the accumulated amount of rain up to including day T
obs ~ [3..30] @ 0 :- state=rainy @ 0.
obs ~ [0..5] @ 0 :- state=sunny @ 0.

obs ~ [R+3..R+30] @ T :-
    state=rainy @ T,
    T > 0,
    obs=R @ T-1.

obs ~ [R..R+5] @ T :-
    state=sunny @ T,
    T > 0,
    obs=R @ T-1.
\end{minted}

The equivalent code in ProbLog is
\begin{minted}{prolog}
0.4::start(weather,sunny); 0.6::start(weather,rainy).
0.6::trans(weather,T,sunny,sunny); 0.4::trans(weather,T,sunny,rainy).
0.3::trans(weather,T,rainy,sunny); 0.7::trans(weather,T,rainy,rainy).

1/6 :: emit(weather,0,sunny, 0);
1/6 :: emit(weather,0,sunny, 1);
1/6 :: emit(weather,0,sunny, 2);
1/6 :: emit(weather,0,sunny, 3);
1/6 :: emit(weather,0,sunny, 4);
1/6 :: emit(weather,0,sunny, 5).


1/ 28 :: emit(weather,0,rainy, 3);
1/ 28 :: emit(weather,0,rainy, 4);
1/ 28 :: emit(weather,0,rainy, 5);
1/ 28 :: emit(weather,0,rainy, 6);
1/ 28 :: emit(weather,0,rainy, 7);
1/ 28 :: emit(weather,0,rainy, 8);
1/ 28 :: emit(weather,0,rainy, 9);
1/ 28 :: emit(weather,0,rainy, 10);
1/ 28 :: emit(weather,0,rainy, 11);
1/ 28 :: emit(weather,0,rainy, 12);
1/ 28 :: emit(weather,0,rainy, 13);
1/ 28 :: emit(weather,0,rainy, 14);
1/ 28 :: emit(weather,0,rainy, 15);
1/ 28 :: emit(weather,0,rainy, 16);
1/ 28 :: emit(weather,0,rainy, 17);
1/ 28 :: emit(weather,0,rainy, 18);
1/ 28 :: emit(weather,0,rainy, 19);
1/ 28 :: emit(weather,0,rainy, 20);
1/ 28 :: emit(weather,0,rainy, 21);
1/ 28 :: emit(weather,0,rainy, 22);
1/ 28 :: emit(weather,0,rainy, 23);
1/ 28 :: emit(weather,0,rainy, 24);
1/ 28 :: emit(weather,0,rainy, 25);
1/ 28 :: emit(weather,0,rainy, 26);
1/ 28 :: emit(weather,0,rainy, 27);
1/ 28 :: emit(weather,0,rainy, 28);
1/ 28 :: emit(weather,0,rainy, 29);
1/ 28 :: emit(weather,0,rainy, 30).

1/6 :: emit(weather, T, sunny, R0); 
1/6 :: emit(weather, T, sunny, R1);
1/6 :: emit(weather, T, sunny, R2);
1/6 :: emit(weather, T, sunny, R3);
1/6 :: emit(weather, T, sunny, R4);
1/6 :: emit(weather, T, sunny, R5) :-
  state(weather, T, sunny),
  T>0,
  TT is T-1,
  observe(weather, TT, R0),
  R1 is R0+1,
  R2 is R0+2,
  R3 is R0+3,
  R4 is R0+4,
  R5 is R0+5.


1/28 :: emit(weather, T, rainy, R1); 
1/28 :: emit(weather, T, rainy, R2);
1/28 :: emit(weather, T, rainy, R3);
1/28 :: emit(weather, T, rainy, R4);
1/28 :: emit(weather, T, rainy, R5);
1/28 :: emit(weather, T, rainy, R6); 
1/28 :: emit(weather, T, rainy, R7);
1/28 :: emit(weather, T, rainy, R8);
1/28 :: emit(weather, T, rainy, R9);
1/28 :: emit(weather, T, rainy, R10);
1/28 :: emit(weather, T, rainy, R11); 
1/28 :: emit(weather, T, rainy, R12);
1/28 :: emit(weather, T, rainy, R13);
1/28 :: emit(weather, T, rainy, R14);
1/28 :: emit(weather, T, rainy, R15);
1/28 :: emit(weather, T, rainy, R16); 
1/28 :: emit(weather, T, rainy, R17);
1/28 :: emit(weather, T, rainy, R18);
1/28 :: emit(weather, T, rainy, R19);
1/28 :: emit(weather, T, rainy, R20);
1/28 :: emit(weather, T, rainy, R21); 
1/28 :: emit(weather, T, rainy, R22);
1/28 :: emit(weather, T, rainy, R23);
1/28 :: emit(weather, T, rainy, R24);
1/28 :: emit(weather, T, rainy, R25);
1/28 :: emit(weather, T, rainy, R26); 
1/28 :: emit(weather, T, rainy, R27);
1/28 :: emit(weather, T, rainy, R28) :-
  state(weather, T, rainy),
  T>0,
  TT is T-1,
  observe(weather, TT, R0),
  R1 is R0 + 3,
  R2 is R0 + 4,
  R3 is R0 + 5,
  R4 is R0 + 6,
  R5 is R0 + 7,
  R6 is R0 + 8,
  R7 is R0 + 9,
  R8 is R0 + 10,
  R9 is R0 + 11,
  R10 is R0 + 12,
  R11 is R0 + 13,
  R12 is R0 + 14,
  R13 is R0 + 15,
  R14 is R0 + 16,
  R15 is R0 + 17,
  R16 is R0 + 18,
  R17 is R0 + 19,
  R18 is R0 + 20,
  R19 is R0 + 21,
  R20 is R0 + 22,
  R21 is R0 + 23,
  R22 is R0 + 24,
  R23 is R0 + 25,
  R24 is R0 + 26,
  R25 is R0 + 27,
  R26 is R0 + 28,
  R27 is R0 + 29,
  R28 is R0 + 30.

%%%%%%%%%
% background: which state are we in at which time?
% state(modelID,time,state)
%%%%%%%%%
state(A,0,S) :- start(A,S).
state(A,T,S) :- T > 0, TT is T-1, state(A,TT,S2), trans(A,TT,S2,S).

%%%%%%%%%
% background: which symbol do we see at which time?
% observe(modelID,time,symbol)
%%%%%%%%%
observe(A,T,S) :- state(A,T,State), emit(A,T,State,S).
\end{minted}

We tested the timing of three sets queries which solve filtering problems: estimating the current state based on past observations. Each set of queries was tested in \Fusemate (with and without grounding optimization) and ProbLog. The following configuration flag can be set to switch on/off grounding optimization by setting \fm{FLAG} to \fm{true}/\fm{false}, respectively, 
\begin{minted}{prolog}
config(query_optimization_grounding, FLAG). 
\end{minted}

\paragraph{All slightly rainy.} The observations at each timepoint increased the amount of water observed by four, a value that can occur from either the sunny or the rainy distributions. The complexity was increased by increasing the number of observations and the timepoint at which the final state was estimated.

In \Fusemate, this was implemented by starting at complexity $N=1$
\begin{minted}{prolog}
config(eot, 1).
?-
    state=S7 @ 1
    |
    obs=4 @ 1.
\end{minted}
and finishing at complexity $N=7$
\begin{minted}{prolog}
config(eot, 7).
?-
    state=S7 @ 7
    |
    obs=4 @ 1,
    obs=8 @ 2,
    obs=12 @ 3,
    obs=16 @ 4,
    obs=20 @ 5,
    obs=24 @ 6,
    obs=28 @ 7.
\end{minted}

The equivalent ProbLog code is to start at $N=1$
\begin{minted}{prolog}
evidence(observe(weather, 1, 4)).
query(state(weather, 1, X)).
\end{minted}
and finish at $N=7$
\begin{minted}{prolog}
evidence(observe(weather, 1, 4)).
evidence(observe(weather, 2, 8)).
evidence(observe(weather, 3, 12)).
evidence(observe(weather, 4, 16)).
evidence(observe(weather, 5, 20)).
evidence(observe(weather, 6, 24)).
evidence(observe(weather, 7, 28)).
query(state(weather, 7, X)).
\end{minted}

\paragraph{All sunny.} Each observation was of zero precipitation, so the observations must have come from the sunny distribution. Complexity $N$ was increased by increasing the number of observations and the timepoint of the final observation.

In \Fusemate, the implementation was to start at $N=1$
\begin{minted}{prolog}
config(eot, 1).
?-
    state=S7 @ 1
    |
    obs=0 @ 1.
\end{minted}
and finish at $N=7$
\begin{minted}{prolog}
config(eot, 7).
?-
    state=S7 @ 7
    |
    obs=0 @ 1,
    obs=0 @ 2,
    obs=0 @ 3,
    obs=0 @ 4,
    obs=0 @ 5,
    obs=0 @ 6,
    obs=0 @ 7.
\end{minted}

Equivalently in ProbLog, starting at $N=1$
\begin{minted}{prolog}
evidence(observe(weather, 1, 0)).
query(state(weather, 1, X)).
\end{minted}
and finish at $N=7$
\begin{minted}{prolog}
evidence(observe(weather, 1, 0)).
evidence(observe(weather, 2, 0)).
evidence(observe(weather, 3, 0)).
evidence(observe(weather, 4, 0)).
evidence(observe(weather, 5, 0)).
evidence(observe(weather, 6, 0)).
evidence(observe(weather, 7, 0)).
query(state(weather, 7, X)).
\end{minted}

\paragraph{Mixed weather.} A variety of observations were included in this query, which meant there was a mixture of observations from both the sunny and rainy distributions. Complexity $N$ was increased by increasing the number of observations and the timepoint at which the final state is predicted.

The \Fusemate implementation starts at complexity $N=1$
\begin{minted}{prolog}
config(eot, 1).
?-
    state=S7 @ 1
    |
    obs=0 @ 1.
\end{minted}
and finishes at $N=7$
\begin{minted}{prolog}
config(eot, 7).
?-
    state=S7 @ 7
    |
    obs=0 @ 1,
    obs=4 @ 2,
    obs=24 @ 3,
    obs=34 @ 4,
    obs=38 @ 5,
    obs=38 @ 6,
    obs=42 @ 7. 
\end{minted}

Equivalently, in ProbLog starting at $N=1$
\begin{minted}{prolog}
evidence(observe(weather, 1, 0)).
query(state(weather, 1, X)).
\end{minted}
and finishing at $N=7$
\begin{minted}{prolog}
evidence(observe(weather, 1, 0)).
evidence(observe(weather, 2, 4)).
evidence(observe(weather, 3, 24)).
evidence(observe(weather, 4, 34)).
evidence(observe(weather, 5, 38)).
evidence(observe(weather, 6, 38)).
evidence(observe(weather, 7, 42)).
query(state(weather, 7, X)).
\end{minted}

\end{document}